\documentclass[letterpaper, 10 pt, conference]{ieeeconf}  
\usepackage{graphicx}
\usepackage{booktabs, multirow, hhline}
\usepackage[table]{xcolor}
\usepackage{amsmath}
\usepackage{tikz}
\usepackage{amssymb}
 \usepackage{pifont}
  
 \usepackage{amssymb}
 \usepackage[ruled,vlined]{algorithm2e}
\usepackage{float}

\usepackage{algorithmic}
\usepackage{xcolor}

\usepackage{amsmath}
\usepackage{tcolorbox}  
\usepackage{bbold}
\usepackage{array}
\usepackage{multirow}
\usepackage{url}
\usepackage{xurl}

\usepackage[colorlinks=true, linkcolor=black, citecolor=blue, urlcolor=blue]{hyperref}

\IEEEoverridecommandlockouts                              

\title{\LARGE \bf
Image-based Geo-localization for Robotics: Are Black-box Vision-Language Models there yet?
}

\author{Sania Waheed$^{1}$, Bruno Ferrarini$^{2}$, Michael Milford $^{3}$,
Sarvapali D. Ramchurn $^{1}$ and Shoaib Ehsan$^{1,4}$
\thanks{This work was supported by the U.K. Engineering and Physical Sciences Research Council under Grant EP/Y009800/1 and Grant EP/V00784X/1}.
\thanks{$^{1}$Sania Waheed and Sarvapali D. Ramchurn are with the School of Electronics and Computer Science, University of Southampton, Southampton SO17 1BJ. {\tt\small (sw1m24@soton.ac.uk; sdr1@soton.ac.uk)}}%
\thanks{$^{2}$Bruno Ferrarini is with MyWay srl, Via Osti, 6 - 29010 Vernasca (PC).{\tt\small (bferrarini.ac.uk@gmail.com)}}%
\thanks{$^{3}$M. Milford is with the School of Electrical Engineering and Computer Science, Queensland University of Technology, Brisbane, QLD 4000, Australia. {\tt\small (michael.milford@qut.edu.au)}}%
\thanks{$^{1,4}$Shoaib Ehsan is with the School of Electronics and Computer Science, University of Southampton, SO17 1BJ Southampton, U.K., and also with the School of Computer Science and Electronic Engineering, University of Essex, CO4 3SQ Colchester, U.K. {\tt\small (s.ehsan@soton.ac.uk)}}%
}
\begin{document}

\maketitle
\thispagestyle{empty}
\pagestyle{empty}



\begin{abstract}
The advances in Vision-Language models (VLMs) offer exciting opportunities for robotic applications involving image geo-localization – the problem of identifying the geo-coordinates of a place based on visual data only. In robotics, such capabilities are particularly relevant to the global re-localization stage of the kidnapped robot problem, where a robot must recover its pose without prior knowledge of its location. Recent work has focused on using a VLM as embedding extractor for geo-localization. However, the most sophisticated VLMs may only be available as black boxes that are accessible through an API, and come with a number of limitations: there is no access to training data, model features and gradients; retraining is not possible; and the number of predictions may be limited by the API. The potential of state-of-the-art VLMs as a stand-alone, zero-shot geo-localization systems at planet scale using a single text-based prompt is largely unexplored. To bridge this gap, this paper undertakes the first systematic study, to the best of our knowledge, to investigate state-of-the-art generative VLMs as stand-alone, zero-shot geo-localization systems in a black-box setting with realistic constraints. We consider three main scenarios for this thorough investigation: a) fixed text-based prompt; b) semantically-equivalent text-based prompts; and c) semantically-equivalent query images. Beyond standard accuracy, we introduce model consistency as a metric to account for the auto-regressive and probabilistic nature of generative VLMs. Our findings reveal that while VLMs demonstrate strong coarse-level localization and navigation priors, fine-grained localization degrades significantly under realistic variations, highlighting reliability challenges for deploying generative VLMs in robust, open-world robotic navigation systems.
\end{abstract}

\section{Introduction}
Vision-Language Models (VLMs) are increasingly being explored for robotic perception and open-world navigation tasks such as autonomous vehicles and robotic systems \cite{shah2022lmnavroboticnavigationlarge, huang2023visuallanguagemapsrobot, dorbala2022clipnavusingclipzeroshot}. A key capability for such systems is accurate, robust, and reliable geo-localization, particularly in scenarios requiring global re-localization, such as the kidnapped robot problem, where a robot must recover its pose without prior knowledge of its location. Image-based geo-localization is particularly valuable in this case because it can provide precise location information in environments where GPS signals are weak, unavailable, or unreliable. However, it remains a highly challenging task due to the vast geographic diversity of our planet and environmental variations such as lighting, weather, and seasonal changes. These challenges make robustness and reliability essential considerations when evaluating geo-localization methods for deployment in real-world applications.

\begin{figure} [tb]
    \centering
    \vspace*{1mm}
    \includegraphics[scale=0.13]{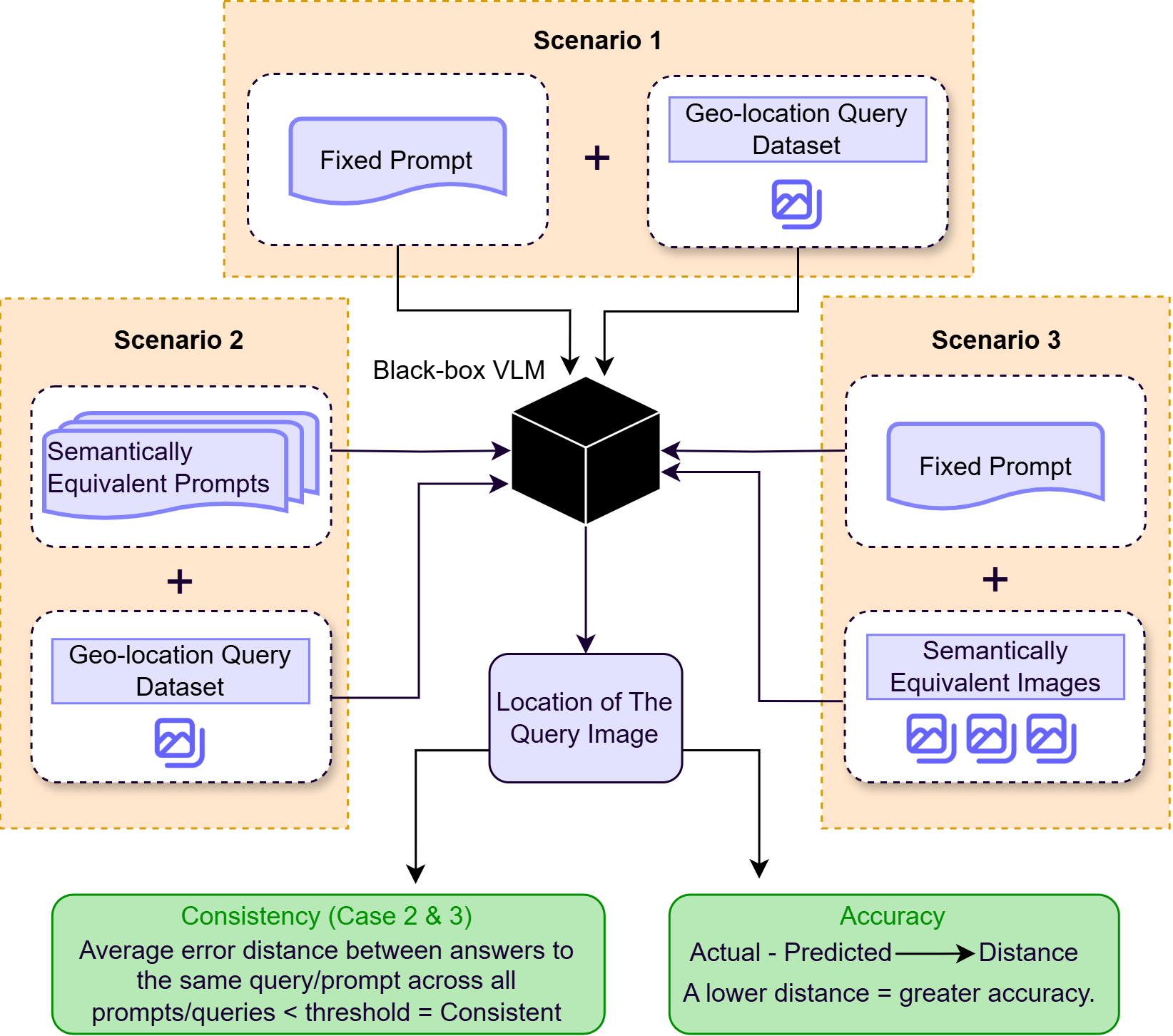}
    \caption{Proposed framework for investigating VLMs in a black-box geo-localization setting. \textbf{Scenario 1}: A fixed prompt is paired with a query image to measure geo-localization accuracy. \textbf{Scenario 2}: Prompt sensitivity is evaluated using semantically-equivalent prompts, measuring accuracy and self-consistency across prompt variations. \textbf{Scenario 3}: Robustness to environmental changes is evaluated using semantically-equivalent images, measuring accuracy and self-consistency.}
    \label{fig:fig1}
\end{figure} 

Traditionally, geo-localization efforts have been focused on classification and retrieval approaches. More recently, VLMs have helped incorporate broader semantic and contextual knowledge into the extracted features. As a result, open-source VLMs have primarily been used as feature extractors \cite{haas2024pigeonpredictingimagegeolocations, 10.1145/3557918.3565868, Zhou_2024} for retrieval. However, many of the most capable proprietary VLMs are only accessible via black-box APIs, limiting direct access to embeddings and restricting their use within conventional retrieval pipelines.

Despite these constraints, recent work \cite{mendes2024granularprivacycontrolgeolocation} demonstrates that SOTA VLMs, such as GPT-4v, are capable geo-locators even in a black-box setting using a single text-based prompt. This emerging capability raises important questions about their reliability and generalization for real-world robotic navigation, where consistent localization performance across diverse environments is essential. Since access to the internal representations of these models is limited, safety assurance and failure analysis becomes more challenging. Furthermore, existing evaluations often rely on publicly available datasets \cite{Hays:2008:im2gps} which have data distributions that may resemble model training data, underscoring the need for a broader, more rigorous assessment that covers a wider data distribution, including both public and non-public datasets. Additionally, sensitivity to semantically-equivalent text-based prompts and semantically-equivalent query images needs to be taken into consideration before deployment of VLMs in geo-localization systems. 

To overcome the above-mentioned shortcomings, this paper performs a systematic study to thoroughly investigate the geo-localization capabilities of SOTA VLMs. To the best of our knowledge, this is the first paper that attempts to answer several key questions in this context: Are black-box VLMs reliable geo-locators using a single text-based prompt? What is the sensitivity of VLMs to semantically-equivalent text-based prompts? What is the sensitivity of VLMs to semantically-equivalent query images? What is the relationship between accuracy and consistency, and how does this affect the reliability of the model?

This paper proposes a new evaluation framework and makes three main contributions (as illustrated in Fig. \ref{fig:fig1}): 
\begin{itemize}
    \item First, we conduct a comprehensive study on the performance of three SOTA VLMs (GPT-4v, IDEFICS-80b-Instruct, and LLaVA-13b) for the geo-localization task in a black-box setting, providing an in-depth analysis of their applicability as stand-alone, zero-shot geo-localization systems. We benchmark the performance of these models on two standard publicly available datasets (IM2GPS \cite{Hays:2008:im2gps} and Tokyo 24/7 \cite{inproceedings}) and two non-public datasets (CSN and LZR \cite{ismagilov2024motionblurdeblurringvisual}). 
    \item Second, we systematically investigate the models' sensitivity to text-based prompt variations (while keeping the query images fixed) in terms of model accuracy and consistency.
    \item Third, we thoroughly examine the models' sensitivity to semantically-equivalent query images with environmental variations (while keeping the text-based prompt fixed), including factors such as lighting, weather, and seasonal changes, highlighting how real-world environmental conditions can significantly influence VLM performance.
\end{itemize}

Our findings reveal that SOTA VLMs exhibit strong performance in black-box geo-localization task at broader spatial levels, such as country or regional level localization ensuring high-level scene understanding. However, their fine-grained localization accuracy does not generalize well, suggesting potential performance inflation due to over representation of certain type of data in the models' training data. Furthermore, these models also struggle to generalize effectively under environmental or prompt variations. 

\begin{figure}[tb]
    \vspace*{1.5mm} 
    \centering
    \includegraphics[scale=0.16]{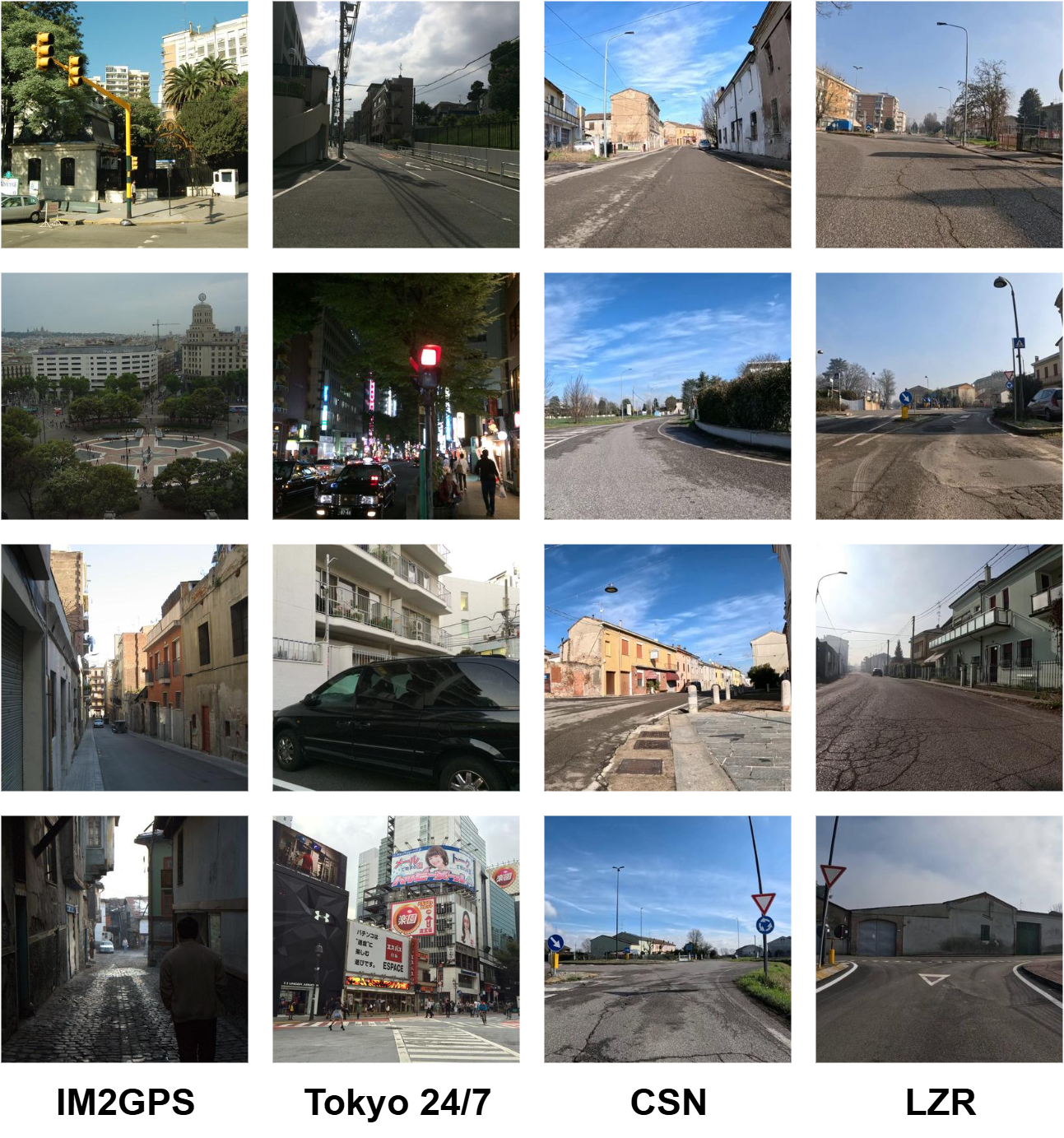}
    \caption{Sample images from the datasets used in our experiments. IM2GPS contains images from diverse locations worldwide, whereas Tokyo 24/7, CSN, and LZR focus on specific regions. Tokyo 24/7 was collected in Tokyo, Japan, while CSN and LZR were captured in the Italian towns.}
    \label{fig:dataset_examples}
\end{figure}

The remainder of this paper is structured as follows: Section II reviews related work, Section III describes the proposed framework and evaluation metrics. Section IV details the experimental setup, including datasets, models, and prompting methods. Section V presents the results and discussions and Section VI concludes the paper with future research directions.

\section{Related Work}
This section provides an overview of the related work in image-based geo-localization and VLMs.
\subsection{Image Based Geo-localization}
This task aims to determine the precise geographic coordinates of a location based solely on visual input. Approaches in this field can be broadly categorized into restricted and planet-scale geo-localization \cite{10.1007, MllerBudack2018GeolocationEO}. Restricted geo-localization focuses on specific environments, such as individual cities or well-known landmarks \cite{kendall2016posenetconvolutionalnetworkrealtime, liu2019stochasticattractionrepulsionembeddinglarge, 8100137, Wang_2020}. In contrast, planet-scale geo-localization tackles the more challenging task of open-world localization and navigation \cite{Hays:2008:im2gps} which is significantly more complex due to the diversity of environments and the variability in image conditions. To address these challenges, geo-localization methods primarily rely on retrieval or classification techniques \cite{8099699, shi2020ilookingatjoint,  workman2015wideareaimagegeolocalizationaerial, zhu2021vigorcrossviewimagegeolocalization,  pramanick2022worldimagetransformerbasedgeolocalization, 10.1007/978-3-030-46147-8_1}. Both these approaches utilize feature extraction models built on CNNs \cite{seo2018cplanetenhancingimagegeolocalization, MllerBudack2018GeolocationEO, kordopatiszilos2021leveragingefficientnetcontrastivelearning} or transformer architectures \cite{pramanick2022worldimagetransformerbasedgeolocalization, zhu2022transgeotransformerneedcrossview}.

\subsection{VLMs in Geo-localization}
VLMs have recently gained prominence in image-based place recognition tasks due to their ability to incorporate semantic and contextual understanding during feature extraction. VLMs, such as CLIP \cite{radford2021learningtransferablevisualmodels} have predominantly been used as feature extractors for image based geo-localization \cite{10.1145/3557918.3565868, luo2022g3geolocationguidebookgrounding, Zhou_2024, haas2024pigeonpredictingimagegeolocations}. Beyond this, VLMs are now being explored for geo-localization task in a black-box setting \cite{mendes2024granularprivacycontrolgeolocation, Zhou_2024, han2024swarmintelligencegeolocalizationmultiagent}.  Notably, GPT-4v \cite{openai2024gpt4technicalreport} has demonstrated exceptional geo-localization accuracy, occasionally outperforming specialized systems. \cite{Zhou_2024} bench-marked GPT-4v using prompting and RAG, while \cite{mendes2024granularprivacycontrolgeolocation} highlighted that even prompt alone is sufficient for capable geo-location. However, much of the current research evaluates on a narrow set of public datasets (IM2GPS \cite{Hays:2008:im2gps}, IM2GPS3k\cite{vo2017revisiting}, YFCC4k\cite{vo2017revisiting}, and YFCC26k\cite{muller2018geolocation}) that share the same underlying data distribution, as they all derive from the YFCC100M dataset \cite{10.1145/2812802} or flickr online photo collection. This raises concerns about performance inflation due to potential overlap with model training data, and leaves open the question of how well these models generalize to different environments. Moreover, the impact of prompt and environmental variations on model predictions also remains unexplored.

\section{Proposed Framework}
This work evaluates SOTA VLMs for geo-localization in a black-box setting across three scenarios (Fig.\ref{fig:fig1}): (i) reliability using a single text-based prompt across datasets with varying distributions, (ii) sensitivity to semantically-equivalent text prompts, and (iii) robustness to environmental variations in query images.

\subsection{Scenarios}
\subsubsection{Scenario 1}
Consistent performance of a model across diverse datasets with different distributions would underscore its true geo-localization capabilities. For this scenario, we adopt the prompt formulation from \cite{mendes2024granularprivacycontrolgeolocation}, applying a least-to-most (LTM) prompting strategy that incrementally addresses sub-tasks to accomplish the main geo-localization objective.\footnote{We use the exact prompt template from \cite{mendes2024granularprivacycontrolgeolocation}, provided in Appendix B.1 of the original paper.} Let \( \mathcal{M} \) denote the model, which takes an image \(x \) a prompt \( P \) as input. The model generates a natural language output \( y \), which contains an explanation and predicted geo-coordinates:
\begin{equation} \label{eq1}
\mathcal{M}(x, P) \rightarrow y.
\end{equation}
From the output \( y \), we extract the predicted geo-coordinates \( p \). The localization error \(\mathcal{L}_{\text{geo}} \) is computed as the haversine distance between the predicted coordinates \( p \) and the ground truth coordinates \( g \).

\subsubsection{Scenario 2}
This scenario evaluates the impact of prompt variations on the geo-localization performance of VLMs using two distinct prompting approaches. For each image \( x\), we apply: (i)  A single-sentence (SS) format, and (ii) A structured, LTM format from \cite{mendes2024granularprivacycontrolgeolocation}.

For each prompting strategy, \( s \in \{\text{SS}, \text{LTM}\} \), we define a set of \( V \) semantically-equivalent prompt variations \( \{P_{s}^j\}_{j=1}^V \). Each prompt \( P_{s}^j \) is paired with the same image \( x\) and passed to the model \( \mathcal{M} \), producing a response \( y_{s}^j \):
\begin{equation} \label{eq2}
\mathcal{M}(x, P_{s}^j) \rightarrow y_{s}^j,
\end{equation}
This allows us to analyze model sensitivity to both, prompting style and phrasing.

\begin{figure}[t]
\vspace{2mm}
\centering
\begin{tcolorbox}[colback=gray!5!white, colframe=gray!75!black, title=Single-Sentence Prompt Variations, width=\linewidth, top=2pt, bottom=2pt, left=4pt, right=4pt]
\small
\begin{itemize}
    \item \textbf{Prompt 1:}What is this location? Give me the geo-coordinates only.
    \item \textbf{Prompt 2: }Where is this place on Earth? Can you predict the geo-coordinates?
    \item \textbf{Prompt 3: }Predict the geo-coordinates for the input image.
    \item \textbf{Prompt 4: }Provide a latitude and longitude for this place.
    \item \textbf{Prompt 5: }Can you specify the exact geo-coordinates for this image?
\end{itemize}
\end{tcolorbox}
\caption{Single-Sentence Prompt Variations}
\label{fig:fig2}
\end{figure}

\subsubsection{Scenario 3}
For this scenario, we evaluate model robustness to real-world environmental changes in the query image while keeping the prompt fixed. We examine the model’s ability to consistently interpret and prioritize the essential features of an image and adhere to the specified task under changing conditions such as illumination, weather, and viewpoint variations. For each location \( i \), we define a set of \( V \) image variations \( \{x_i^j\}_{j=1}^V \). Each image \( x_i^j \) is paired with the same LTM prompt \( P \), producing a response \( y_i^j \):
\begin{equation} \label{eq3}
\mathcal{M}(x_i^j, P) \rightarrow y_i^j.
\end{equation}

\subsection{Evaluation Metrics}
\subsubsection{Accuracy}
Let $p = (p_{lat}, p_{lon})$ and $g = (g_{lat}, g_{lon})$ denote the predicted and ground truth coordinates for a given image. The error distance \textit{d} is computed using the Haversine distance:
\begin{equation} \label{eq4}
d(p, g) = Haversine(p_{\text{lat}}, p_{\text{lon}}, g_{\text{lat}}, g_{\text{lon}})
\end{equation}
Accuracy is the proportion of predictions having \textit{d} within threshold \( T \), where \( \mathbb{1} \)  is an indicator function and N is the total number of images. 
\begin{equation} \label{eq5}
Accuracy = \frac{1}{N} \sum_{i=1}^{N} \mathbb{1}(d(p_i, g_i) < T)
\end{equation}

\subsubsection{Consistency}
\begin{figure*}[ht]
\vspace{5mm} 
\centering
\includegraphics[scale=0.14]{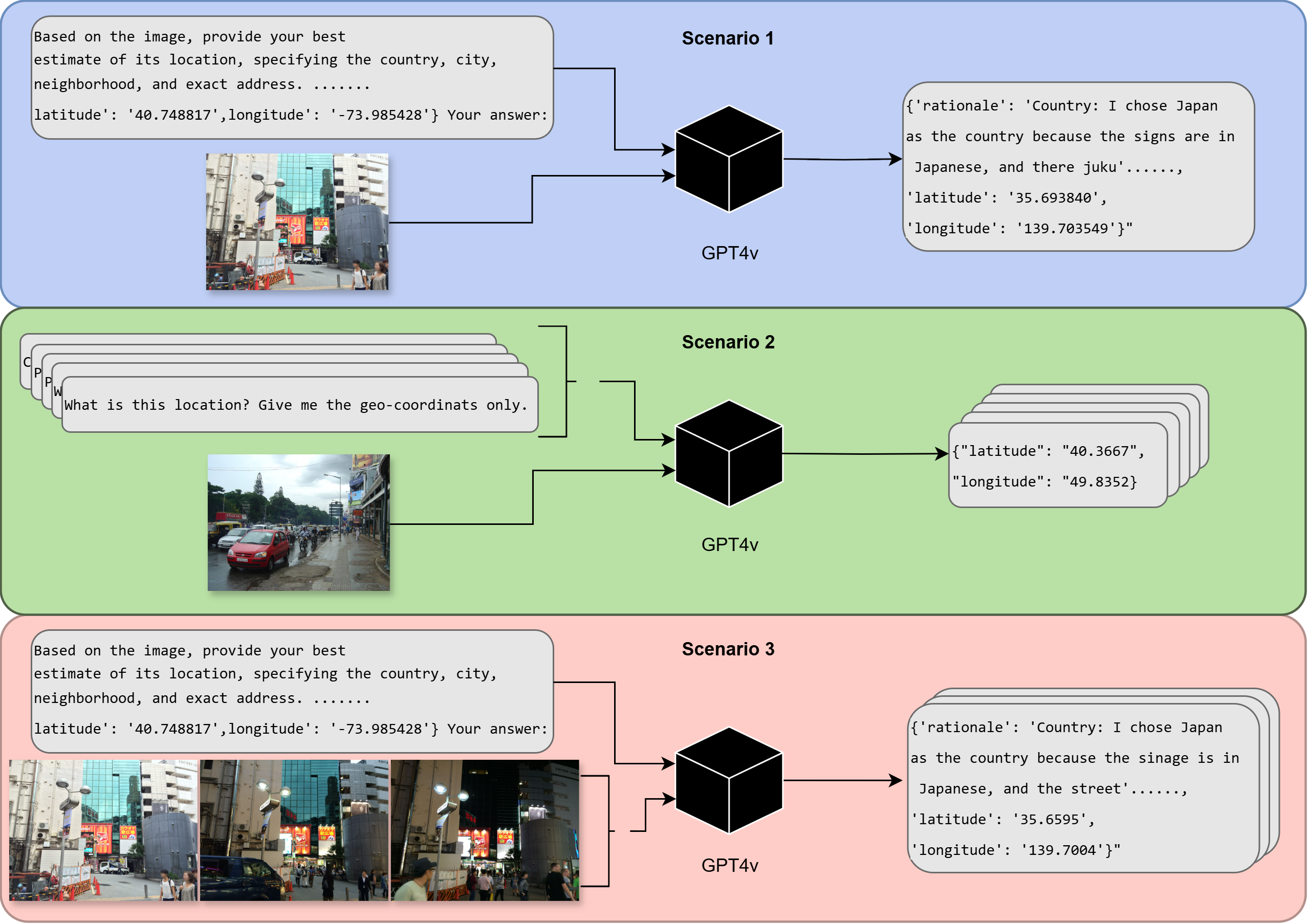} 
\caption{Examples using GPT-4v across the three proposed scenarios.}
\label{fig:methodology_examples}
\end{figure*}

Consistency measures the stability of a generative VLM's outputs across semantically-equivalent inputs, focusing on agreement between predictions irrespective of ground truth. We use this as an additional evaluation metric for Scenarios 2 and 3.

Let \( p_i^j \) denote the predicted coordinates for the \textit{j-th} variation of a location or prompt \( i \), where \( j, k \in \{1, \dots, V\} \) 
index the \( V \) variations (e.g., day, sunset, night for images or semantically-equivalent text for prompts) and \( j \neq k \). The pairwise consistency indicator is:
\begin{equation} \label{eq6}
   C_i^{jk} = 
   \begin{cases} 
      1, & \text{if } d(p_i^j, p_i^k) \leq T, \\
      0, & \text{otherwise.}
   \end{cases}
\end{equation}
The consistency for each location or prompt \( i \) is averaged over all unique pairs:
\begin{equation} \label{eq7}
C_i = \frac{2}{V(V-1)} \sum_{j=1}^{V-1} \sum_{k=j+1}^{V} C_i^{jk},
\end{equation}
where \( \frac{2}{V(V-1)} \) is the normalization factor that accounts for the total number of unique pairs and the overall model consistency is:
\begin{equation} \label{eq8}
\textit{Model Consistency} = \frac{1}{M} \sum_{i=1}^{M} C_i,
\end{equation}
where \( M \) is the total number of locations or prompts in the dataset.

\section{Experimental Setup}
This section provides the details of the datasets and models used, the prompting strategies employed, and the specific evaluation metrics applied in our experiments. 


\subsection{Datasets}
The IM2GPS test set \cite{Hays:2008:im2gps}, a standard geo-localization benchmark derived from flickr, has a single-image-per-location structure, which restricts evaluation under varying conditions. To address this, we include three additional datasets: Tokyo 24/7 \cite{inproceedings} (315 images captured across daytime, sunset, and nighttime) introducing viewpoint and time-of-day variations; and two non-public datasets, CSN and LZR \cite{ismagilov2024motionblurdeblurringvisual}, captured along the same outdoor routes under varying weather, illumination, and viewpoint conditions in urban and countryside environments in Casoni (CSN) and Luzzara (LZR), Italy. The data collection protocol, acquisition setup, and ethical considerations for CSN and LZR are detailed in \cite{ismagilov2024motionblurdeblurringvisual}. For evaluation, we use 100 corresponding images per traverse from morning and noon traverses in CSN, and morning and dusk traverses in LZR. Fig. \ref{fig:dataset_examples}, presents samples from all four datasets.
\subsection{Models}
We evaluate three state-of-the-art VLMs in our experiments: GPT-4v (gpt-4o-2024-05-13)\cite{openai2024gpt4technicalreport}, IDEFICS-80b-Instruct \cite{OBELICS}, and LLaVA-1.5-13b \cite{liu2024improvedbaselinesvisualinstruction}. These models were selected to represent diverse model paradigms (closed-source, open-weight, and open-source), and scales, offering a comparison at varying levels of model accessibility. We also conducted preliminary tests with smaller variants such as IDEFICS-9b-Instruct and LLaVA-1.5-7b, but they showed negligible geo-localization performance, achieving less than 1\% accuracy even at a 2500 km threshold.
\begin{figure}[t]
    \centering
    \includegraphics[width=0.95\columnwidth]{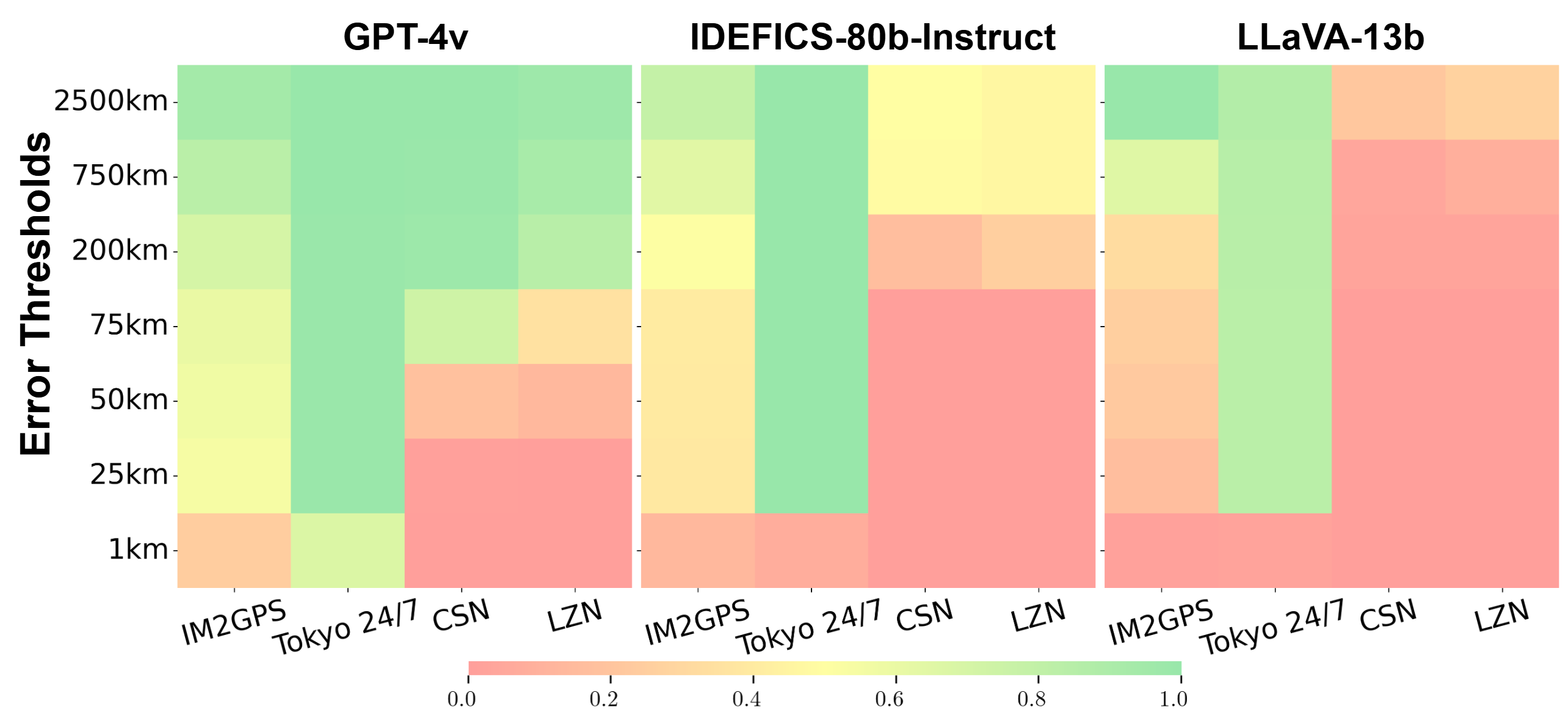}
    \caption{Geo-localization performance of GPT-4v, IDEFICS-80b-Instruct, and LLaVA-13b across datasets and localization error thresholds (1–2500 km). The results highlight model generalization across distinct data distributions, including public and private datasets.}
    \label{fig:Scenario1-Fig2}
\end{figure}
\vspace{-3mm}

\subsection{Prompting}
All three scenarios use the LTM prompting method from \cite{mendes2024granularprivacycontrolgeolocation}\footnote{The full prompt is omitted for brevity; we refer the reader to \cite{mendes2024granularprivacycontrolgeolocation} for details.}. For Scenario 2, we additionally evaluate five single-sentence prompt variations (Fig. \ref{fig:fig2}) alongside five semantically-equivalent LTM variations adapted from \cite{mendes2024granularprivacycontrolgeolocation}; the latter are omitted here for brevity.

\subsection{Evaluation}
The predicted geo-coordinates are extracted from model responses using the regular expression \texttt{r'[-+]?\textbackslash d*\textbackslash .\textbackslash d+|\textbackslash d+'}. Evaluation follows standard geo-localization thresholds for both accuracy and consistency: Street (1 km), City (25 km), Region (200 km), Country (750 km), and Continent (2500 km), providing insights into the model’s reliability across different geographic scales. For instances where models fail to generate geo-coordinates, the error distance is set to $\infty$ to ensure such cases are accounted for during evaluation.

\section{Results and Discussions}

In this section, we present the results of our experiments across three scenarios. The findings highlight the impact of various factors on geo-localization performance, including data distribution, prompt formulation and variations, and environmental conditions, on model accuracy and consistency. An example of the three proposed scenarios is shown in Fig. \ref{fig:methodology_examples} using GPT-4v.
\begin{figure}[h]
    \centering
    \includegraphics[scale=0.13]{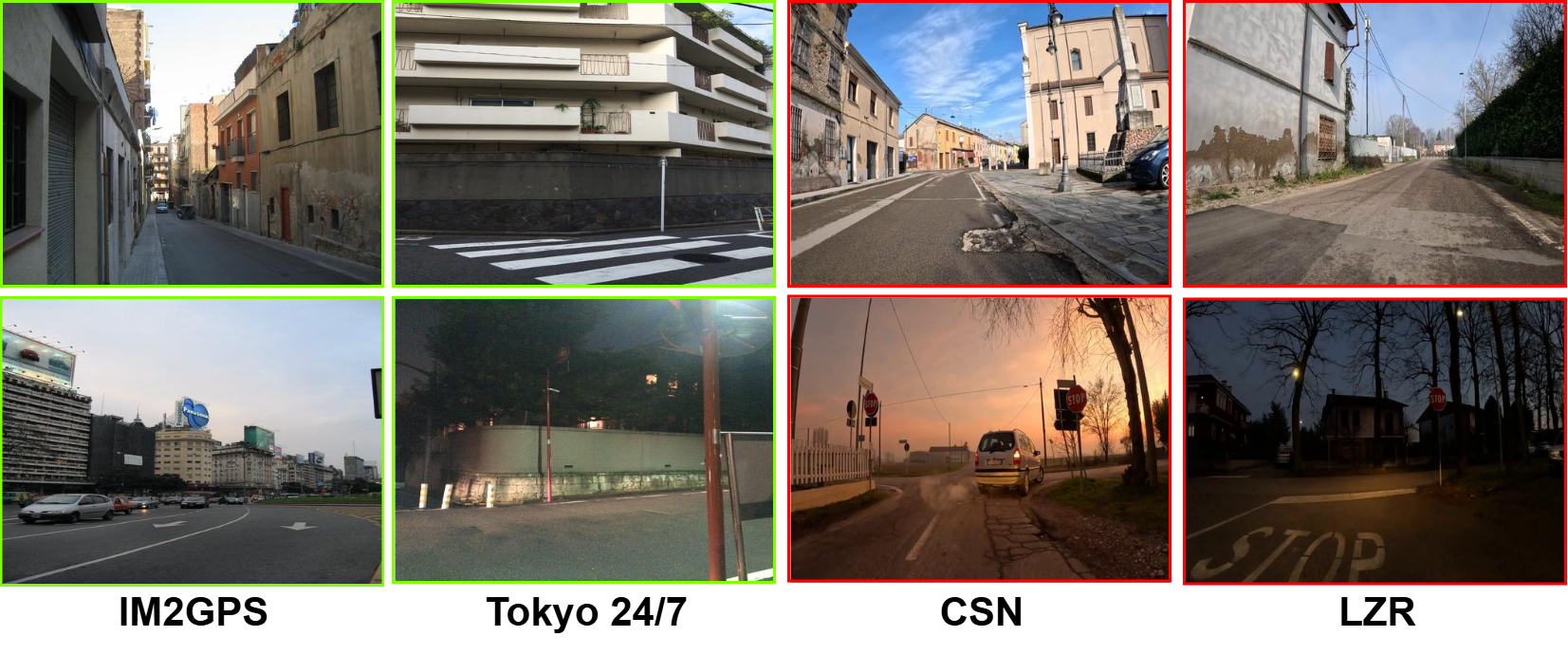}
    \caption{Comparison of localization performance at the 1km threshold for public (IM2GPS, Tokyo 24/7) and non-public (CSN, LZR) datasets. Each row shows examples with similar visual characteristics (e.g., roads, buildings, or signs), where public dataset images are correctly localized while images from non-public datasets fail. This highlights the difficulty VLMs face in generalizing to unseen locations at fine-grained levels.}
    \label{fig:1km_comparison}
\end{figure}
\vspace{-3mm}
\subsection{Scenario 1: Evaluation of Geo-Localization Generalization}
Fig.\ref{fig:Scenario1-Fig2} presents heat-maps illustrating the performance of GPT-4v, IDEFICS-80b-Instruct, and LLaVA-13b across various datasets and error thresholds. Each heatmap highlights performance trends across models and datasets at different geographic scales.

As seen in Fig.\ref{fig:Scenario1-Fig2}, all three models perform significantly better on web scraped or public datasets, IM2GPS and Tokyo 24/7, compared to the non-public datasets, CSN and LZR. This discrepancy likely stems from the models training data distribution overleaping with the public data. GPT-4v achieving 70\% street-level accuracy for Tokyo 24/7 highlights the models' potential for geo-localization in urban and familiar settings. However, the significant drop in performance with the non-public datasets underscores challenges in generalizing to unfamiliar locations. Fig.\ref{fig:1km_comparison} further supports this observation by showing visually similar images across all four datasets where despite similar visual content only the public dataset images are accurately localized within 1km. This further suggests that the performance gap stems not only from visual ambiguity, but may also result from the models' exposure to similar data during training.

Interestingly, the largest accuracy gains for the non-public datasets occur between the 50km and 200km thresholds. This behavior reflects the inherent structure of the datasets, where the images share common visual elements that the models successfully use to make region- or country-level predictions. Beyond the 200km threshold, accuracy becomes saturated. While this demonstrates VLMs' capability to generalize at coarser scales, it also highlights their limitations as fine-grained geo-locators.

\subsection{Scenario 2: Robustness to Prompt Strategy and Variations}
\begin{figure}[t]
    \centering
    \vspace*{1mm}
    \includegraphics[width=0.95\columnwidth]{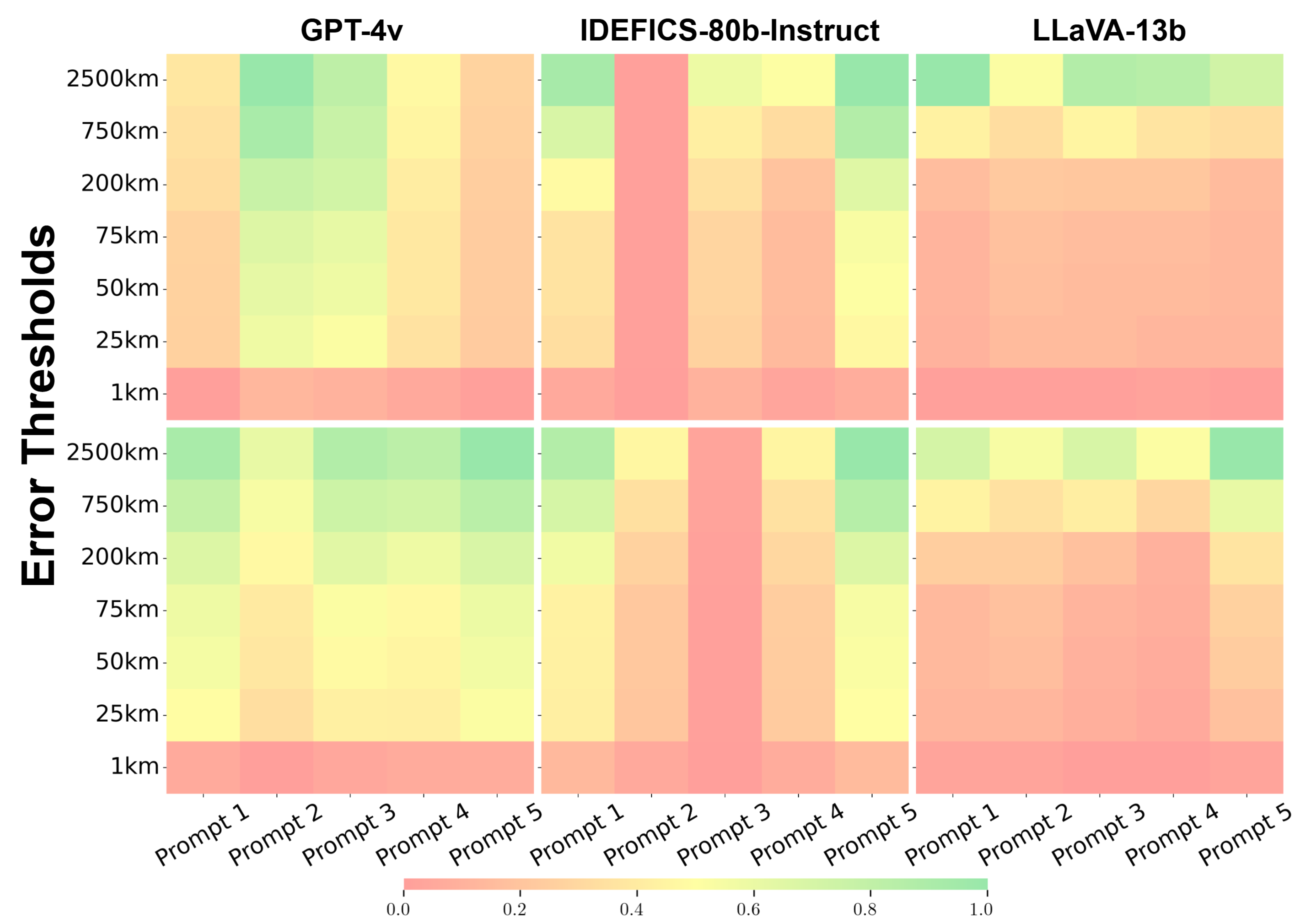}
    \caption{Impact of prompt formulation on geo-localization performance. \textbf{Single-sentence} prompts (top row) exhibit significant variability and lower accuracy, while structured \textbf{LTM} prompts (bottom row) improve overall performance and consistency. However, even with LTM prompts, performance remains sensitive to semantically equivalent variations in phrasing.}
    \label{fig:Scenario 2 - Fig 1}
\end{figure}

The results in Fig.~\ref{fig:Scenario 2 - Fig 1} highlight the significant impact of prompt formulation on geo-localization accuracy across all three models on the IM2GPS dataset. Single-sentence prompts consistently underperform compared to the LTM prompt, emphasizing the importance of richer contextual cues for spatial reasoning. The single-sentence prompts also exhibit high variability, with no formulation achieving consistent performance across all models. In contrast, the LTM prompt reduces this instability and shows clearer performance trends.

During consistency evaluation, we observed that some models, particularly LLaVA-13b, occasionally replicate the geo-coordinates present in the prompt instead of generating their own predictions. This behavior artificially inflates the consistency scores because repeated outputs appear consistent despite not reflecting genuine reasoning. To mitigate this issue, such instances are excluded before computing the final consistency scores. Table~\ref{tab:tab1} shows that LTM prompting improves consistency for both GPT-4s and IDEFICS-80b-Instruct by eliciting more stable reasoning patterns. However, consistency and accuracy do not always align. For example, although IDEFICS-80b-Instruct achieves higher accuracy than LLaVA-13b across most prompt variants in both single-sentence and LTM formulations, its predictions remain highly inconsistent, resulting in lower consistency scores. This issue is particularly pronounced for single-sentence prompts, where IDEFICS-80b-Instruct’s consistency drops below that of LLaVA-13b.

\begin{table}[t]
    \centering
    \caption{Consistency for single-sentence prompt and LTM prompt on IM2GPS dataset.}
    \renewcommand{\arraystretch}{1.05} 
    \resizebox{\columnwidth}{!}{
    \begin{tabular}{c|cc|cc|cc}
        \toprule
        \multirow{2}{*}{Threshold (km)} & \multicolumn{2}{c|}{GPT-4v} & \multicolumn{2}{c|}{IDEFICS-80b-Instruct} & \multicolumn{2}{c}{LLaVA-13b} \\
        \cmidrule{2-7}
        & Single & LTM & Single & LTM & Single & LTM \\
        \midrule
        1    & 0.254 & \textbf{0.328} & 0.003 & \textbf{0.009} & \textbf{0.003} & 0.000 \\
        25   & 0.303 & \textbf{0.453} & 0.000 & \textbf{0.018} & \textbf{0.017} & 0.006 \\
        50   & 0.307 & \textbf{0.472} & 0.005 & \textbf{0.019} & \textbf{0.022} & 0.006 \\
        75   & 0.309 & \textbf{0.482} & 0.008 & \textbf{0.019} & \textbf{0.026} & 0.007 \\
        100  & 0.310 & \textbf{0.486} & 0.008 & \textbf{0.019} & \textbf{0.030} & 0.007 \\
        200  & 0.320 & \textbf{0.503} & 0.008 & \textbf{0.020} & \textbf{0.035} & 0.008 \\
        750  & 0.327 & \textbf{0.541} & 0.017 & \textbf{0.030} & 0.054 & \textbf{0.189} \\
        2500 & 0.329 & \textbf{0.576} & 0.060 & \textbf{0.061} & 0.079 & \textbf{0.315} \\
        \bottomrule
    \end{tabular}
    }
    \label{tab:tab1}
\end{table}

\vspace{-3mm}

\begin{table}[h]
    \centering
    \vspace{2mm}
    \definecolor{pastelgreen}{RGB}{204, 255, 204}
    \definecolor{pastelred}{RGB}{255, 204, 204}
    \renewcommand{\arraystretch}{1.05} 
    \setlength{\tabcolsep}{2pt} 
    \caption{Each dataset group includes images captured under different lighting conditions (e.g., Tokyo 24/7$_{Day}$, $_{Sunset}$, $_{Night}$). Comparisons are made within each dataset group.}
    \resizebox{\columnwidth}{!}{ 
    \begin{tabular}{@{}lcccccccc@{}}
        \toprule
        \textbf{Dataset} & \textbf{1 km} & \textbf{25 km} & \textbf{50 km} & \textbf{75 km} & \textbf{100 km} & \textbf{200 km} & \textbf{750 km} & \textbf{2500 km} \\
        \midrule
        \multicolumn{9}{c}{\textbf{GPT-4v}} \\
        \midrule
        Tokyo 24/7$_{Day}$       & \cellcolor{pastelred}65.38 & \cellcolor{pastelred}97.11 & \cellcolor{pastelred}97.11 & \cellcolor{pastelred}97.11 & \cellcolor{pastelred}97.11 & \cellcolor{pastelred}97.11 & \cellcolor{pastelred}99.03 & \cellcolor{pastelred}99.03 \\
        Tokyo 24/7$_{Sunset}$    & \cellcolor{pastelred}67.61 & \cellcolor{pastelgreen}100 & \cellcolor{pastelgreen}100.00 & \cellcolor{pastelgreen}100.00 & \cellcolor{pastelgreen}100.00 & \cellcolor{pastelgreen}100.00 & \cellcolor{pastelgreen}100.00 & \cellcolor{pastelgreen}100.00 \\
        Tokyo 24/7$_{Night}$     & \cellcolor{pastelgreen}69.52 & \cellcolor{pastelgreen}100 & \cellcolor{pastelgreen}100.00 & \cellcolor{pastelgreen}100.00 & \cellcolor{pastelgreen}100.00 & \cellcolor{pastelgreen}100.00 & \cellcolor{pastelgreen}100.00 & \cellcolor{pastelgreen}100.00 \\
        CSN$_{Morning}$     & 0 & \cellcolor{pastelred}0 & \cellcolor{pastelred}15.00 & \cellcolor{pastelgreen}75.00 & \cellcolor{pastelgreen}81.00 & \cellcolor{pastelgreen}98.00 & 99.00 & 100.00 \\
        CSN$_{Noon}$     & 0 & \cellcolor{pastelgreen}2 & \cellcolor{pastelgreen}21.00 & \cellcolor{pastelred}72.00 & \cellcolor{pastelred}77.00 & \cellcolor{pastelred}97.00 & 99.00 & 100.00 \\
        LZR$_{Morning}$   & 0 & 0 & \cellcolor{pastelgreen}20.00 & \cellcolor{pastelgreen}44.00 & \cellcolor{pastelgreen}56.00 & \cellcolor{pastelgreen}97.00 & \cellcolor{pastelgreen}99.00 & \cellcolor{pastelgreen}100.00 \\
        LZR$_{Dusk}$     & 0 & 0 & \cellcolor{pastelred}7.00 & \cellcolor{pastelred}26.00 & \cellcolor{pastelred}31.00 & \cellcolor{pastelred}71.00 & \cellcolor{pastelred}85.00 & \cellcolor{pastelred}94.00 \\
        \midrule
        \multicolumn{9}{c}{\textbf{IDEFICS-80b-Instruct}} \\
        \midrule
        Tokyo 24/7$_{Day}$       & \cellcolor{pastelgreen}8.5 & \cellcolor{pastelred}71.42 & \cellcolor{pastelred}71.42 & \cellcolor{pastelred}71.42 & \cellcolor{pastelred}71.42 & \cellcolor{pastelred}71.42 & \cellcolor{pastelred}71.42 & \cellcolor{pastelred}71.42 \\
        Tokyo 24/7$_{Sunset}$    & \cellcolor{pastelred}4.7 & \cellcolor{pastelgreen}76.19 & \cellcolor{pastelgreen}76.19 & \cellcolor{pastelgreen}76.19 & \cellcolor{pastelgreen}76.19 & \cellcolor{pastelgreen}76.19 & \cellcolor{pastelgreen}76.19 & \cellcolor{pastelgreen}76.19 \\
        Tokyo 24/7$_{Night}$     & \cellcolor{pastelred}3.80 & \cellcolor{pastelred}73.33 & \cellcolor{pastelred}73.33 & \cellcolor{pastelred}73.33 & \cellcolor{pastelred}73.33 & \cellcolor{pastelred}73.33 & \cellcolor{pastelred}73.33 & \cellcolor{pastelred}73.33 \\
        CSN$_{Morning}$   & 0 & 0 & 0 & 0 & 0 & \cellcolor{pastelgreen}16.00 & \cellcolor{pastelgreen}50.00 & \cellcolor{pastelgreen}52.00 \\
        CSN$_{Noon}$     & 0 & 0 & 0 & 0 & 0 & \cellcolor{pastelred}8.00 & \cellcolor{pastelred}21.00 & \cellcolor{pastelred}22.00 \\
        LZR$_{Morning}$   & 0 & 0 & 0 & 0 & 0 & \cellcolor{pastelgreen}36.00 & \cellcolor{pastelgreen}59.00 & \cellcolor{pastelgreen}59.00 \\
        LZR$_{Dusk}$      & 0 & 0 & 0 & 0 & 0 & \cellcolor{pastelred}1.00 & \cellcolor{pastelred}9.00 & \cellcolor{pastelred}9.00 \\
        \midrule
        \multicolumn{9}{c}{\textbf{LLaVA-13b}} \\
        \midrule
        Tokyo 24/7$_{Day}$       & \cellcolor{pastelred}0.95 & \cellcolor{pastelgreen}30.47 & \cellcolor{pastelgreen}30.47 & \cellcolor{pastelgreen}30.47 & \cellcolor{pastelgreen}30.47 & \cellcolor{pastelgreen}31.42 & \cellcolor{pastelgreen}31.42 & \cellcolor{pastelgreen}32.38 \\
        Tokyo 24/7$_{Sunset}$    & \cellcolor{pastelgreen}1.90 & \cellcolor{pastelred}27.61 & \cellcolor{pastelred}27.61 & \cellcolor{pastelred}27.61 & \cellcolor{pastelred}27.61 & \cellcolor{pastelred}27.61 & \cellcolor{pastelred}28.57 & \cellcolor{pastelred}28.57 \\
        Tokyo 24/7$_{Night}$     & \cellcolor{pastelred}0 & \cellcolor{pastelgreen}30.47 & \cellcolor{pastelgreen}30.47 & \cellcolor{pastelgreen}30.47 & \cellcolor{pastelred}30.47 & \cellcolor{pastelred}30.47 & \cellcolor{pastelred}30.47 & \cellcolor{pastelred}31.42 \\
        CSN$_{Morning}$   & 0 & 0 & 0 & 0 & 0 & 1.00 & \cellcolor{pastelred}1.00 & \cellcolor{pastelgreen}9.00 \\
        CSN$_{Noon}$      & 0 & 0 & 0 & 0 & 0 & 1.00 & \cellcolor{pastelgreen}2.00 & \cellcolor{pastelred}6.00 \\
        LZR$_{Morning}$  & 0 & 0 & 0 & 0 & 0 & 1.00 & \cellcolor{pastelgreen}5.00 & \cellcolor{pastelgreen}15.00 \\
        LZR$_{Dusk}$      & 0 & 0 & 0 & 0 & 0 & 1.00 & \cellcolor{pastelred}1.00 & \cellcolor{pastelred}4.00 \\
        \bottomrule
    \end{tabular}
    }
    \label{tab:tab2}
\end{table}

\subsection{Scenario 3: Robustness to Environmental Variations}
The accuracy results in Table \ref{tab:tab2} show the strong influence of environmental conditions on VLM geo-localization. Across datasets, models generally perform worse in darker conditions than in well-lit ones. This trend is most pronounced in the non-public CSN and LZR datasets, where morning settings often outperform dusk or noon by up to 50\%. These findings underscore the importance of accounting for environmental conditions, a factor often overlooked in prior work. In contrast, the Tokyo 24/7 dataset shows stable performance across daytime, sunset, and nighttime, with differences typically within 3–4\%. This is likely because images in Tokyo 24/7 dataset contains many textual cues (e.g., street signs, shop names, billboards) that remain legible regardless of time-of-day, providing strong semantic information that reduces environmental impact.

The consistency results in Table \ref{tab:tab3} reveal further patterns across models and datasets. GPT-4v is highly consistent across datasets but experiences a notable decline on LZR, likely due to its severe lighting transitions from bright daytime to dark dusk. CSN’s milder shifts (daytime to noon) lead to better consistency. The disparity in model consistency between public and non-public datasets supports the possibility of training data contamination. Interestingly, the consistency of both GPT-4v and IDEFICS-80b-Instruct stabilizes at broader thresholds (e.g., beyond 100 km), suggesting that environmental conditions mainly affect fine-grained localization, while city- or country-scale predictions remain stable. LLaVA-13b consistently struggles across all datasets in terms of both accuracy and consistency, highlighting its poor performance due to its significantly smaller size compared to the other two models.


\begin{table}[t]
    \centering
    \caption{Consistency of \textbf{GPT-4v}, \textbf{IDEFICS-80b-Instruct}, and \textbf{LLaVA-13b} under environmental variations. Performance degrades under environmental changes (e.g., LZR daytime-to-dusk transitions) at fine-grained thresholds, but stabilizes at coarser scales.}
    \renewcommand{\arraystretch}{1.05} 
    \resizebox{\columnwidth}{!}{
    \begin{tabular}{c|ccc|ccc|ccc}
        \toprule
        \multirow{2}{*}{\textbf{Threshold (km)}} & \multicolumn{3}{c|}{\textbf{GPT-4v}} & \multicolumn{3}{c|}{\textbf{IDEFICS-80b-Instruct}} & \multicolumn{3}{c}{\textbf{LLaVA-13b}} \\
        \cmidrule{2-10}
        & \textbf{Tokyo 24/7} & \textbf{CSN} & \textbf{LZR} & \textbf{Tokyo 24/7} & \textbf{CSN} & \textbf{LZR} & \textbf{Tokyo 24/7} & \textbf{CSN} & \textbf{LZR} \\
        \midrule
        1    & 0.61 & 0.04 & 0.02 & 0.38 & 0.37 & 0.30 & 0.00 & 0.05 & 0.06 \\
        25   & 0.98 & 0.34 & 0.17 & 0.65 & 0.37 & 0.30 & 0.06 & 0.07 & 0.06\\
        50   & 0.98 & 0.42 & 0.20 & 0.65 & 0.37 & 0.30 & 0.06 & 0.07 & 0.06\\
        75   & 0.98 & 0.48 & 0.20 & 0.65 & 0.37 & 0.30 & 0.06 & 0.07 & 0.06\\
        100  & 0.98 & 0.62 & 0.22 & 0.65 & 0.37 & 0.30 & 0.06 & 0.07 & 0.06\\
        200  & 0.98 & 0.89 & 0.55 & 0.65 & 0.37 & 0.30 & 0.07 & 0.07 & 0.06\\
        750  & 0.99 & 0.99 & 0.84 & 0.65 & 0.42 & 0.30 & 0.07 & 0.07 & 0.07\\
        2500 & 0.99 & 1 & 0.94 & 0.65 & 0.43 & 0.30 & 0.08 & 0.08 & 0.08\\ 
        \bottomrule
    \end{tabular}
    }
    \label{tab:tab3}
\end{table}

\section{Conclusions and Future Work}
This work presented a systematic analysis of SOTA VLMs for zero-shot image-based geo-localization in a black-box setting. Our findings show that while these models demonstrate promising coarse-level localization capabilities, fine-grained performance degrades significantly under prompt and environmental variations, particularly in unseen environments. Furthermore, localization accuracy and prediction consistency are not always correlated, suggesting that accuracy alone may overestimate reliability for robotic applications. Overall, current black-box VLMs appear better suited for providing high-level semantic or regional localization priors rather than serving as standalone fine-grained localization systems, especially under practical robotic constraints such as real-time inference, onboard compute limitations, and robustness requirements. This study is limited to three representative VLMs and does not include newer frontier-scale multimodal models. Future work should extend this benchmark to newer models, investigate hybrid localization approaches, and further address ethical considerations such as privacy and unintended location disclosure from publicly shared imagery.









\bibliographystyle{IEEEtran}
\bibliography{IEEEexample}

\end{document}